\documentclass{article}
\usepackage{spconf,amsmath,epsfig}
\usepackage{graphicx} % Required for inserting images
\usepackage[utf8]{inputenc} % allow utf-8 input
\usepackage[T1]{fontenc}    % use 8-bit T1 fonts
\usepackage{hyperref}
\hypersetup{
    colorlinks=true,
    linkcolor=blue,
    filecolor=magenta,      
    urlcolor=cyan,
    citecolor = blue,
    pdftitle={Causality for food insecurity},
    pdfpagemode=FullScreen,
    }
\usepackage{cite}   
\usepackage{url}            % simple URL typesetting
\usepackage{booktabs}       % professional-quality tables
\usepackage{amsfonts}       % blackboard math symbols
\usepackage{nicefrac}       % compact symbols for 1/2, etc.
\usepackage{microtype}      % microtypography
\usepackage[square,numbers]{natbib}
\usepackage{multirow}
\usepackage{comment}

\title{Assessing the Causal Impact of Humanitarian Aid on Food Security}
\name{Jordi Cerdà-Bautista, \hspace{0.5cm} José María Tárraga, \hspace{0.5cm} Vasileios Sitokonstantinou, \hspace{0.5cm} Gustau Camps-Valls}
\address{Image Processing Laboratory\\	Universitat de València\\ \{jordi.cerda, jose.maria.tarraga, vasileios.sitokonstantinou, gustau.camps\}@uv.es}

\begin{document}
%\ninept
%
\maketitle
\begin{abstract}
In the face of climate change-induced droughts, vulnerable regions encounter severe threats to food security, demanding urgent humanitarian assistance. This paper introduces a causal inference framework for the Horn of Africa, aiming to assess the impact of cash-based interventions on food crises. Our contributions include identifying causal relationships within the %food security
malnutrition system, harmonizing a comprehensive database including socio-economic, weather and remote sensing data, and estimating the causal effect of %humanitarian
cash-based interventions on malnutrition. On a country level, our results revealed no significant effects, likely due to limited sample size, suboptimal data quality, and an imperfect causal graph resulting from our limited understanding of multidisciplinary systems like %food security.
malnutrition. Instead, on a district level, results revealed significant effects, further implying the context-specific nature of the system. This underscores the need to enhance data collection and refine causal models with domain experts for more effective future interventions and policies, improving transparency and accountability in humanitarian aid.
\end{abstract}
\section{Introduction}
\label{sec:intro}

In a world where climate change is rapidly accelerating, droughts are becoming more frequent and severe, posing a serious challenge to food security in the most vulnerable regions of our planet. In this context, communities that rely solely on rainfall for their livelihoods are especially at risk, often requiring immediate humanitarian assistance to survive \citep{Funk2018, Pape2019}. Failure to act or provide adequate aid can have immense consequences, including devastating economic losses, mass displacement of people, malnutrition in infants, and elevated mortality rates due to hunger and famine \citep{Desai2021, SomaliaDisplaced2022, Maxwell2011}. Humanitarian organizations are facing a significant challenge due to the widening gap between funding and the needs of the people affected by food crises \citep{WFP2023, FAO2023}. As a result, designing effective humanitarian interventions in resource-constrained situations has become a critical issue. Despite numerous comprehensive reviews, there is still a lack of solid evidence to identify the best strategies to help populations affected by crises \citep{Shannon2017}. Cash-based and voucher aid programs are considered effective in emergencies, but their cost-effectiveness varies by context \citep{Shannon2018}. Standardized methods for evaluating %humanitarian
cash-based interventions in food emergencies are lacking \citep{Shannon2017}. Our aim is to determine the impact of interventions, using observational causal inference to enhance intervention design, and transparency in charity, and improve %humanitarian
cash-based aid outcomes during extreme droughts.

The Horn of Africa has witnessed a concerning rise in acute malnutrition, affecting 6.5 million people in 2022 \citep{WFP2023}. Prolonged dry spells significantly contribute to this crisis \citep{Coughlan2019}, yet it is crucial to recognize that droughts are not the sole driver. Various factors, including hydrological conditions, food production capabilities, market access, insufficient %humanitarian
aid, conflicts, and displacement, play a significant role \citep{Maxwell2023, Sneyers2017, GFDRR2017, Warsame2023, Abel2019}. Studying %food security 
malnutrition in this context is intricate, involving multiple variables, scales, and non-linear relationships. Predictive Machine Learning (ML) techniques are not suited to understanding the causes and estimating the causal effect by default \citep{Pearl2000, Peters2017}, instead, this paper focuses on causal inference, specifically assessing the impact of %humanitarian
cash-based interventions during the 2016, 2018, and 2022 Horn of Africa droughts. Our aim is to demonstrate the application of causal inference for evaluating the effectiveness of cash-based interventions in %food crisis 
malnutrition scenarios.

\section{Related Work}
\label{sec:related}

In recent years, the surge in available data has enabled us to assess the impact of climate change on food insecurity. This data originates from diverse sources, encompassing Earth observation products \citep{CHIRPS} and systematic socioeconomic data collection programs \citep{FSNAU, PRMN}. Leveraging this wealth of data, we can estimate causal effects from observations \citep{Pearl2000, Peters2017}. This approach is particularly vital in domains where conducting controlled experiments is impractical, costly, or unethical, with food insecurity research being a prominent example. Observational data for causal inference has gained prominence across various disciplines, including ecology \citep{Arif2022}, agriculture \citep{giannarakis2022towards, tsoumas2023evaluating, deines2019satellites}, public policy \citep{Fougere2019, Fougere2021}, and Earth sciences \citep{Runge2021, Perez2018}. While there have been subjective and technical assessments of %humanitarian
cash-based interventions in emergency contexts \citep{Shannon2017, HumResponseReview2017}, to the best of our knowledge, this is the first effort to apply modern observational causal inference methods to evaluate %humanitarian policy
the effectiveness of cash-based aid in a food emergency context. It is also the first time such a broad database of driving factors has been used for this purpose. The contributions of our work are summarized as follows: i) identifying the overarching causal graph and the drivers of %food insecurity
malnutrition in the Horn of Africa, ii) building a harmonized database with the best available data suitable to evaluate cash-based interventions, iii) the estimation of the causal effect of cash-based interventions on malnutrition.

\section{Causal Inference}

%Machine Learning (ML) techniques have achieved considerable success in various domains, including computer vision, natural language processing, graph representation learning, and reinforcement learning. However, deploying these models in real-world scenarios presents challenges such as reduced generalization performance with shifts in data distribution \citep{Damour2020}, a lack of control over generative model samples \citep{Jahanian2020}, biased predictions perpetuating unfair discrimination \citep{Mehrabi2021, Bender2021}, abstract interpretability notions \citep{Lipton2018}, and difficulties in translating reinforcement learning to practical applications \citep{Dulac2019}. These issues are partially attributed to the absence of causal formalism in modern ML systems, leading to a growing interest in causal machine learning (CausalML), which incorporates causal knowledge into ML methods \citep{Kaddour2022}.

Deep Learning (DL) techniques have achieved considerable success in various domains, including computer vision, natural language processing or graph representation learning. DL has shown increased evidence of the potential to address problems in Earth and climate sciences as well \citep{Reichstein2019}. Since 2014, DL applied to Earth Observation has grown exponentially, triggered by the extensive and highly available data sources, and the methodological advancements in DL \citep{Camps2021}. However, deploying DL models in real-world scenarios presents challenges such as reduced generalization performance with shifts in data distribution \citep{Damour2020}, biased predictions perpetuating unfair discrimination \citep{Mehrabi2021, Bender2021}, or abstract interpretability notions \citep{Lipton2018}. These issues are partially attributed to the absence of causal formalism in modern Machine Learning (ML) systems, leading to a growing interest in causal machine learning (CausalML), which incorporates causal knowledge into ML methods \citep{Kaddour2022}.

To reason about the causal effects of certain random variables on others, first, we need to codify causal relations. Causal inference provides a language for formalizing structural knowledge about the data-generating process \citep{pearl_2009} with which we can estimate what will happen to data after changes, called interventions. The canonical representation of causal relations is a causal Direct Acyclic Graph (DAG), which can encode a priori assumptions about the causal structure of interest. In causal modeling, assumptions are crucial, as establishing relations solely on observational data can prove challenging. Background knowledge and domain expertise are a common source of assumptions in causal inference. Randomized controlled trials (RCTs) are also a gold standard for establishing causality because, under certain conditions, random assignment helps control for confounding variables. However, RCTs also bring to attention when this sort of experiment cannot be performed. When studying complex dynamic systems such as the %food security
malnutrition system, replicating interventional experiments could prove infeasible or unethical. Therefore, when modeling the causal relations of the %food security
malnutrition system involving climate and socio-economic dynamics, multi-scale and non-linear drivers, we rely solely on the information provided by background knowledge and associated literature.

\section{Data and Methods}
\label{sec:data_methods}

\subsection{Notation and terminology.} 

\begin{figure}[h]
\includegraphics[width=8 cm]{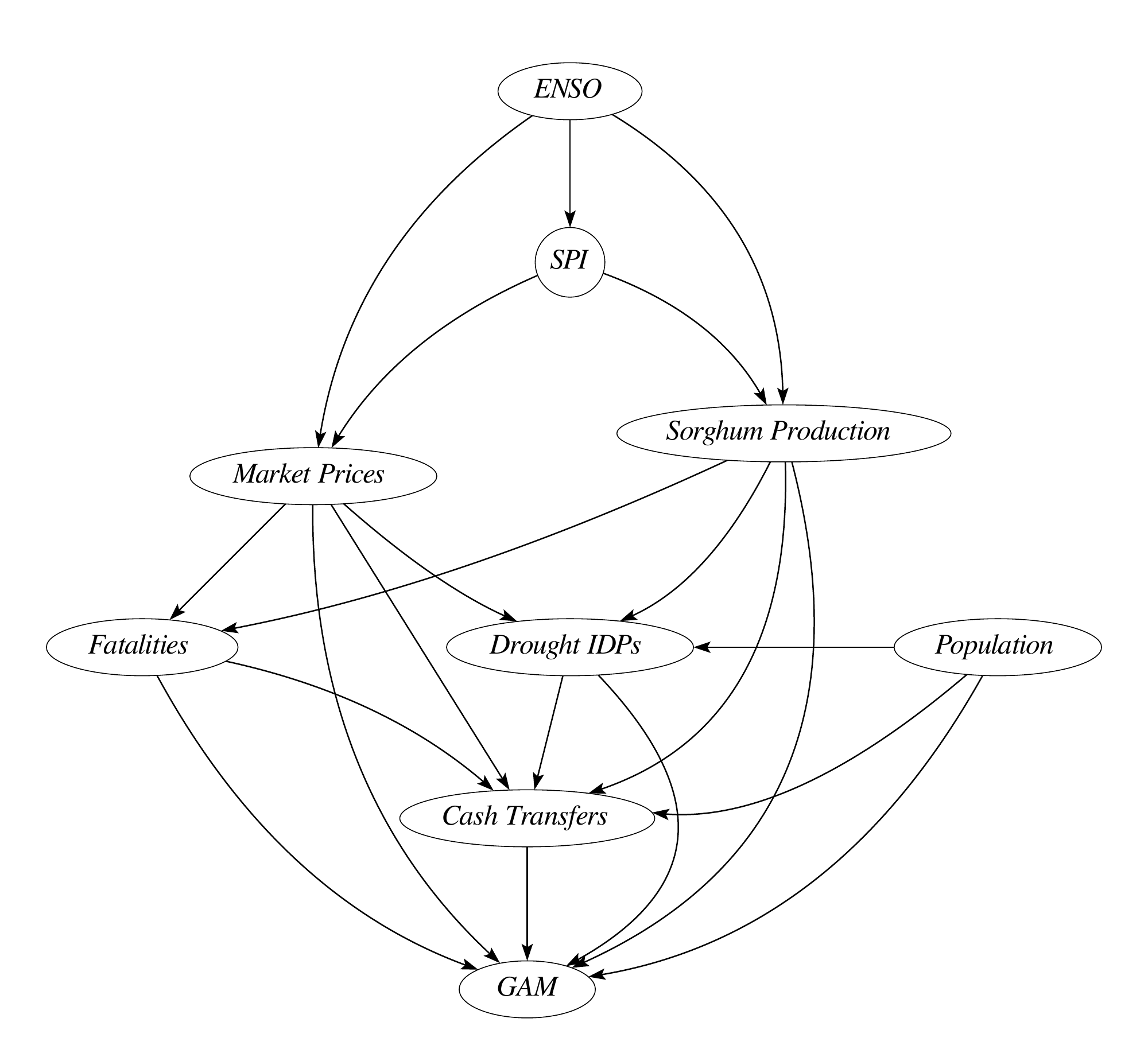}
\caption{DAG representing the %food security
malnutrition system in Somalia.}
\label{fig:causal_graph}
\end{figure}

In this paper, we assess the impact of cash interventions (treatment) on malnutrition (outcome) using a DAG denoted as $G \equiv (V,E)$ (see Figure \ref{fig:causal_graph}). The set of vertices, labeled by $V$, represents relevant variables, and directed edges in set $E$ indicate causation from one variable to another \citep{pearl_2009}. We employ the $do$-operator to describe interventions. $\mathbb{P}(Y=y|do(T=t))$ denotes the probability that $Y=y$ when we intervene by setting the value of $T$ to $t$. Here, $T$ is the treatment variable (cash interventions), and $Y$ is the outcome variable (Global Acute Malnutrition, GAM).

\begin{table*}[t]
\centering
\caption{{\bf Variables and sources used in the study, temporal and spatial resolution.}}
\begin{tabular}{llll}
\hline
Variable & Source & Spatial Resolution & Temporal Resolution \\
\hline
ENSO & \href{https://climexp.knmi.nl/selectdailyindex.cgi?id=someone@somewhere}{WMO}~\citep{WMO} & Country & Daily \\
Standarized precipitation Index &  \href{https://developers.google.com/earth-engine/datasets/catalog/UCSB-CHG_CHIRPS_DAILY#bands}{CHIRPS}~\citep{CHIRPS} & $0.05$º & Daily \\
Violent Conflict & \href{https://acleddata.com/#/dashboard}{ACLED}~\citep{ACLED} & Geolocated Event & Hourly\\
Local Market Prices & \href{https://fsnau.org/ids/dashboard.php}{FSNAU}~\citep{FSNAU} & District & Monthly \\
Sorghum Yield Production & \href{https://fsnau.org/ids/dashboard.php}{FSNAU}~\citep{FSNAU} & District & Seasonal \\
Drought Displacement & \href{https://unhcr.github.io/dataviz-somalia-prmn/index.html}{UNHCR PRMN}~\citep{PRMN} & District & Weekly\\
Somalia Districts & \href{https://data.humdata.org/dataset/cod-ab-som}{UNDP}~\citep{UNDP} & District &  Static\\
Population & \href{https://data.humdata.org/dataset/cod-ps-som?}{UNFPA}~\citep{UNFPA} & District & Static, 2021 \\
Number of individuals that received cash & \href{https://fsnau.org/ids/dashboard.php}{FSNAU}~\citep{FSNAU} & District & Monthly \\
Global Acute Malnutrition & \href{https://fsnau.org/ids/dashboard.php}{FSNAU}~\citep{FSNAU} & District & Monthly \\

\hline
\end{tabular}
\label{tab:data}
\end{table*}

\subsection{Data} 

%Food security
Malnutrition is influenced by various climatic, economic, and social factors, as represented in our DAG, which reflects the dynamics of agropastoralist households in drought displacement situations in Somalia \citep{idmc}. We collect and harmonize data for the variables in the DAG from multiple sources (Table \ref{tab:data}). Our outcome variable is the %food security
malnutrition index GAM~\citep{FSNAU}, and the treatment variable is a proxy for cash interventions, reflecting the number of individuals who received money in the form of credit or remittances~\citep{FSNAU}. We also collect data on  El Niño Southern Oscillation (ENSO) to account for climate variability~\citep{Sazib} and use the Standardized Precipitation Index (SPI) to characterize dry spells~\citep{SPI2012, CHIRPS}. Socio-economic data include monthly market prices of livestock, staple food, water, and sorghum production~\citep{Acted2017}. We measure conflict levels using a proxy based on recorded fatalities~\citep{thalheimer2023} and incorporate data on drought-induced internal displacement \citep{PRMN}. All data are aggregated annually and by district.

\subsection{Problem Formulation}

%We approach the task of assessing the impact of humanitarian aid on reducing food insecurity as an Average Treatment Effect (ATE) estimation challenge. 

To estimate the Average Treatment Effect (ATE), $\text{ATE} = \mathop{\mathbb{E}}[Y|do(T=1)] - \mathop{\mathbb{E}}[Y|do(T=0)]$, we identify an adjustment set $Z \subseteq V$. We apply the back-door criterion, which relies on a graphical test to determine whether adjusting for a set of graph nodes $Z \subseteq V$ is sufficient for estimating $\mathbb{P}(Y=y|do(T=t))$. We find the parent adjustment set that is sufficient for estimating the ATE, which is $Z=$\{Market Prices, Sorghum Production, Fatalities, Drought-induced internal displacements, Population\}. Utilizing the Potential Outcomes framework, our ATE estimation aims to capture the difference between the average GAM values under %humanitarian
cash-based aid exceeding a chosen threshold and the average value of the outcome when %humanitarian
cash-based aid falls below that threshold. To estimate the effect, we use several methods of varying complexity. Linear regression (LR) and distance matching (M) are selected as baseline estimation methods. The popular Inverse Propensity Score weighting (IPS W) is also used \citep{stuart}, as well as modern machine learning methods, the T-learner (T-L) and X-learner (X-L) \citep{kunzel}. Given the unavailability of observed ground truth estimates, we resort to performing refutation tests, in line with recent research \citep{sharma2020dowhy, cinelli2019}, to assess the robustness of our models. We perform the following tests: i) Placebo treatment, where the treatment is randomly permuted, and the estimated effect is expected to drop to 0; ii) Random Common Cause (RCC), where a random confounder is added to the dataset and the estimate is expected to remain unchanged; iii) Random Subset Removal (RSR), where a subset of data is randomly removed and the effect is expected to remain the same.

\begin{table*}[h]
    \centering
    \caption{{\bf Area, threshold, method, ATE, 95\% confidence intervals and p-values. Refutation tests fail if their p-value is less than 0.05. Numbers are in the percentage of people in GAM per capita. Results for Somalia and the Baidoa district.}}
    \small
    \begin{tabular}{cccccccccccc}
    \hline
        \multicolumn{6}{c}{\multirow{2}{*}{Cause Effect Estimation}} & \multicolumn{6}{c}{Refutation Tests} \\ \cline{7-12} 
        \multicolumn{6}{c}{} & \multicolumn{2}{c}{Placebo} & \multicolumn{2}{c}{RCC} & \multicolumn{2}{c}{RSR} \\ \hline
        Area & Th & Method & ATE & CI & p-value & Effect* & p-value & Effect* & p-value & Effect* & p-value \\
        & & ($10^{-4}$) &  &  & ($10^{-4}$) & & ($10^{-4}$) &  & ($10^{-4}$) & \\ \hline
         & 35 & LR & -0.743 & (-0.001, 0.001) & 0.842 & -0.000 & 0.860 & -0.000 & 0.880 & -0.000 & 0.940 \\ 
         & 35 & M & -3.917 & (-0.001, 0.000) & 0.306 & 0.000 & 0.940 & -0.000 & 1.000 & -0.000 & 0.900 \\ 
         & 35 & IPS W & -1.170 & (-0.001, 0.001) & 0.787 & 1.056 & 0.680 & -1.17 & 1.000 & -1.310 & 1.000 \\ 
         & 35 & T-L (RF) & -1.348 & (-0.001, 0.000) & 0.342 & -1.348 & 0.000 & -1.490 & 0.300 & -1.875 & 0.380 \\ 
         & 35 & X-L (RF) & -2.335 & (-0.001, 0.000) & 0.342 & -1.348 & 0.000 & -2.346 & 0.450 & -2.726 & 0.410 \\ 
         & 50 & LR & -0.168 & (-0.001, 0.001) & 0.971 & 0.000 & 0.960 & -0.000 & 0.980 & -0.000 & 0.960 \\ 
         & 50 & M & -3.725 & (-0.001, 0.000) & 0.380 & -0.000 & 0.880 & -0.000 & 1.000 & -0.000 & 0.740 \\ 
        Somalia & 50 & IPS W & -2.844 & (-0.001, 0.001) & 0.554 & -0.615 & 0.860 & -2.844 & 1.000 & -2.769 & 0.940 \\ 
        (Country) & 50 & T-L (RF) & -0.875 & (-0.001, 0.000) & 0.254 & -0.875 & 0.000 & -1.351 & 0.110 & -2.177 & 0.320 \\ 
         & 50 & X-L (RF) & -2.318 & (-0.001, 0.000) & 0.254 & -0.875 & 0.000 & -2.335 & 0.480 & -2.880 & 0.410 \\ 
         & 75 & LR & -2.139 & (-0.001, 0.001) & 0.690 & -0.000 & 0.980 & -0.000 & 0.900 & -0.000 & 0.980 \\ 
         & 75 & M & 1.970 & (-0.001, 0.001) & 0.750 & -0.000 & 0.840 & 0.000 & 1.000 & 0.000 & 0.720 \\ 
         & 75 & IPS W & -3.508 & (-0.001, 0.000) & 0.509 & -3.964 & 0.440 & -3.508 & 1.000 & -3.868 & 0.920 \\ 
         & 75 & T-L (RF) & -2.631 & (-0.001, 0.000) & 0.095 & -2.631 & 0.000 & -2.584 & 0.480 & -3.211 & 0.420 \\ 
         & 75 & X-L (RF) & -3.630 & (-0.001, 0.000) & 0.095 & -2.631 & 0.000 & -3.577 & 0.460 & -4.415 & 0.350 \\ \hline
         & 35 & T-L (RF) & -7.545 & (-0.001, -0.000) & 0.001 & -7.545 & 0.000 & -8.153 & 0.141 & -9.282 & 0.296 \\ 
         & 35 & X-L (RF) & -1.583 & (-0.001, 0.000) & 0.001 & -7.545 & 0.000 & -2.750 & 0.104 & -5.213 & 0.105 \\ 
         & \textbf{50} & \textbf{M} & \textbf{-15.197} & \textbf{(-0.003, -0.000)} & \textbf{0.025} & \textbf{-0.000} & \textbf{0.980} & \textbf{-0.002} & \textbf{1.000} & \textbf{-0.002} & \textbf{0.960} \\ 
         & \textbf{50} & \textbf{IPS W} & \textbf{-16.968} & \textbf{(-0.003, -0.001)} & \textbf{0.003} & \textbf{-4.833} & \textbf{0.520} & \textbf{-16.968} & \textbf{1.000} & \textbf{-16.886} & \textbf{0.980} \\ 
        Baidoa & 50 & T-L (RF) & -9.377 & (-0.002, -0.000) & 0.022 & -9.377 & 0.000 & -10.260 & 0.075 & -11.317 & 0.261 \\ 
        (District) & 50 & X-L (RF) & -4.816 & (-0.001, 0.000) & 0.022 & -9.377 & 0.000 & -5.750 & 0.100 & -7.610 & 0.213 \\ 
         & \textbf{75} & \textbf{IPS W} & \textbf{-15.898} & \textbf{(-0.003, -0.000)} & \textbf{0.040} & \textbf{-0.632} & \textbf{0.960} & \textbf{-15.898} & \textbf{1.000} & \textbf{-15.801} & \textbf{0.980} \\
         & 75 & T-L (RF) & -9.802 & (-0.002, -0.000) & 0.019 & -9.802 & 0.000 & -10.104 & 0.391 & -11.583 & 0.314 \\ 
         & 75 & X-L (RF) & -3.374 & (-0.001, 0.000) & 0.019 & -9.802 & 0.000 & -4.200 & 0.250 & -7.025 & 0.241 \\ \hline
    \end{tabular}
    \label{tab:results}
\end{table*}

\section{Implementation and Results}
\label{sec:results}

For the experiments, we are using the doWhy \citep{sharma2020dowhy} and Causal ML \citep{chen2020causalml} Python libraries. As the treatment is a continuous variable, we binarize it assuming different thresholds. This can be interpreted as considering the treated group as those samples where the number of individuals who receive money surpass a certain threshold and the control group as the rest. We also remove samples where the treatment is close to the threshold in order not to violate the stable unit treatment value assumption (SUTVA) \citep{pearl_2009}. We take the 35, 50, and 75 percentile values of the number of people who receive any form of cash as different threshold levels.

From 2016 to 2022, we collected data spanning 57 districts in Somalia, resulting in a dataset of 378 samples. To address population differences between urban and agro-pastoral areas, we normalized the data per district population. We framed the problem as an ATE estimation task by converting the number of people receiving money into a binary variable using various thresholds. The estimation represents the percentage of malnourished people who would have been affected if specific thresholds of people receiving money had been reached (Table \ref{tab:results}). While all estimations show a reduction in the percentage of people with GAM as more individuals receive cash interventions, none are statistically significant at the 95\% confidence level. This outcome is expected due to data scarcity and the complexity of the real problem. It is impossible to account for all system drivers, but ongoing efforts aim to improve our understanding and reduce bias by addressing unaccounted major drivers and acquiring more observational data. The humanitarian community has established data repositories, but there's a need for enhanced and broader data collection following FAIR principles (Findability, Accessibility, Interoperability, and Reusable). Additionally, our country-level DAG may not fully capture context-specific relationships and localized impacts on the ground, including factors like past drought events, the political situation, poverty levels, and livelihood options, which significantly influence intervention effectiveness \citep{idmc}.

We perform a last experiment, where we only consider a single district, Baidoa, as we know it contains above-average data quality. We resample the time resolution to monthly to increase the sample size, even though it doesn't provide additional information for seasonal variables, and run the same experiments as in the country-level case. We find statistically significant results in most of the experiments, reaffirming the conclusion that both data quality and localization of the problem are key features of these experiments, and causal assumptions are not fulfilled if these aspects are not considered.

\section{CauseMe platform }
\label{sec:causeme}

Given the data quality challenges and the context-driven nature of %food security
malnutrition systems, our focus turns to data-driven causal discovery as a crucial methodology. The complexity of multi-domain issues, such as %food security
malnutrition, renders traditional expertise insufficient for constructing robust DAGs. Enter the innovative CauseMe platform \citep{Causeme}, functioning as the link between causal inference experts and domain specialists. This platform empowers non-causality experts to employ data-driven causal discovery methods, facilitating data exploration and initial DAG construction. %These structures can then be refined and enhanced by the insights of domain specialists, enabling a comprehensive depiction and understanding of the intricate causal relationships within such complex systems.
Developed as an accessible tool for causal discovery methods, the platform \citep{Runge19} aims to democratize access to these techniques across diverse scientific fields. Catering to experts less versed in causality but eager to conduct further data experiments, CauseMe allows the execution of various causal discovery methods on time series data through an interactive interface. Users can tweak method parameters, select variables for causal analysis, and obtain graphical representations showcasing the returned causal relationships. Moreover, CauseMe facilitates the interpretation of these graphs through a Large Language Model (LLM), allowing users to input contextual information about the data, thereby receiving explanations %aligning the obtained graph links with the dataset's context.
of the obtained graph.

\section{Conclusions}
\label{sec:conclusions}

Optimally distributing available resources and evaluating how, who, where, and when to intervene is crucial to mitigating climate change impacts. In this paper, we presented a novel data-driven approach for assessing the effectiveness of %humanitarian
cash-based interventions in food emergencies through the lens of causal inference. We constructed a DAG to capture the dynamics of %food insecurity
malnutrition under drought conditions and collected data characterizing the system. Our goal was to estimate the causal effects of cash-based interventions on reducing district-level %food insecurity
global acute malnutrition across Somalia. Preliminary country-wise results did not reach statistical significance, although a singular district analysis did, prompting further steps: i) identifying more suitable treatment variables, ii) refining the causal graph with domain experts, iii) gaining insights on the spatio-temporal heterogeneity of impact of interventions through Conditional Average Treatment Effects (CATE) \citep{giannarakis2022personalizing}. If data allows it, causal inference can be used to assess the efficacy of interventions in specific locations, supporting targeted aid where on-ground surveys are not feasible. The proposed approach could promote greater accountability and transparency amongst humanitarian actors, encouraging individuals to contribute to impactful and traceable aid. 

\section{Acknowledgments}

This work was supported by the Fundación BBVA with the project \href{https://www.fbbva.es/noticias/concedidas-5-ayudas-a-equipos-de-investigacion-cientifica-en-big-data/}{`Causal inference in the human biosphere coupled system (SCALE)'}, the Microsoft Climate Research Initiative through the \href{https://www.microsoft.com/en-us/research/collaboration/microsoft-climate-research-initiative/projects/}{Causal4Africa} project, the European Union's Horizon Europe Research and Innovation Program through the \href{https://www.euspa.europa.eu/thinkingearth-copernicus-foundation-models-thinking-earth}{ThinkingEarth} project (under Grant Agreement number 101130544) and the GVA \href{https://isp.uv.es/ai4cs}{PROMETEO} AI4CS
project on ‘AI for complex systems’ (2022-2026) with
CIPROM/2021/056.

% -------------------------------------------------------------------------

\bibliographystyle{IEEEbib}
\bibliography{strings,refs}

\end{document}